\documentclass[letterpaper, 10 pt, conference]{ieeeconf}  % Comment this line out if you need a4paper

\IEEEoverridecommandlockouts                              % This command is only needed if 
\usepackage{graphics} % for pdf, bitmapped graphics files
\usepackage{graphicx}
\usepackage{epsfig} % for postscript graphics files
\usepackage{amsmath} % assumes amsmath package installed
\usepackage{amssymb}  % assumes amsmath package installed
\usepackage{subfigure}
\usepackage{float}
\usepackage{url}
\usepackage{multirow}
\usepackage{subfigure}
\usepackage{wrapfig}
\usepackage{tabularx}
\usepackage{tablefootnote}
\usepackage{hyperref}
\usepackage[dvipsnames]{xcolor}
\usepackage{bm}
\usepackage{array}
\usepackage{booktabs}
\graphicspath{{./figs/}}

\newcolumntype{C}[1]{>{\centering\let\newline\\\arraybackslash\hspace{0pt}}m{#1}}

\newcommand{\doctitle}{Driven to Distraction: Self-Supervised Distractor Learning for Robust Monocular Visual Odometry in Urban Environments}

\newcommand{\docsubtitle}{}

\title{\LARGE \bf\doctitle\docsubtitle} 
\author{Dan Barnes, Will Maddern, Geoffrey Pascoe and Ingmar Posner 
\thanks{Authors are from the Oxford Robotics Institute, Dept. Engineering Science, University of Oxford, UK. 
{\tt\small \{dbarnes,wm,gmp,ingmar\}@robots.ox.ac.uk}}
}

\begin{document}
\maketitle
\thispagestyle{empty}
\pagestyle{empty}

%%%%%%%%%%%%%%%%%%%%%%%%%%%%%%%%%%%%%%%%%%%%%%%%%%%%%%%%%%%%%%%%%%%%%%%%%%%%%%%%
\begin{abstract}

We present a self-supervised approach to ignoring ``distractors'' in camera images for the purposes of robustly estimating vehicle motion in cluttered urban environments.
We leverage offline multi-session mapping approaches to automatically generate a per-pixel \emph{ephemerality mask} and depth map for each input image, which we use to train a deep convolutional network.
At run-time we use the predicted ephemerality and depth as an input to a monocular visual odometry (VO) pipeline, using either sparse features or dense photometric matching.
Our approach yields metric-scale VO using only a single camera and can recover the correct egomotion even when 90\% of the image is obscured by dynamic, independently moving objects.
We evaluate our robust VO methods on more than 400km of driving from the Oxford RobotCar Dataset and demonstrate reduced odometry drift and significantly improved egomotion estimation in the presence of large moving vehicles in urban traffic.
% The project video can be found at 
% \footnotemark.
% \vspace{-2mm}

\end{abstract}

%%%%%%%%%%%%%%%%%%%%%%%%%%%%%%%%%%%%%%%%%%%%%%%%%%%%%%%%%%%%%%%%%%%%%%%%%%%%%%%%

\section{Introduction}\label{sec:introduction}

Autonomous vehicle operation in crowded urban environments presents a number of key challenges to any system based on visual navigation and motion estimation.
In urban traffic where up to 90\% of an image can be obscured by a large moving object (e.g. bus or truck), standard outlier rejection schemes such as RANSAC \cite{fischler1981random} will produce incorrect motion estimates due to the large consensus of features tracked on the moving object. 
The key to robust ``distraction-free'' visual navigation is a deeper understanding of which image regions are \emph{static} and which are \emph{ephemeral} in order to better decide which features to use for motion estimation.

\begin{figure}[]
  \centering
  \includegraphics[width=1.0\columnwidth]{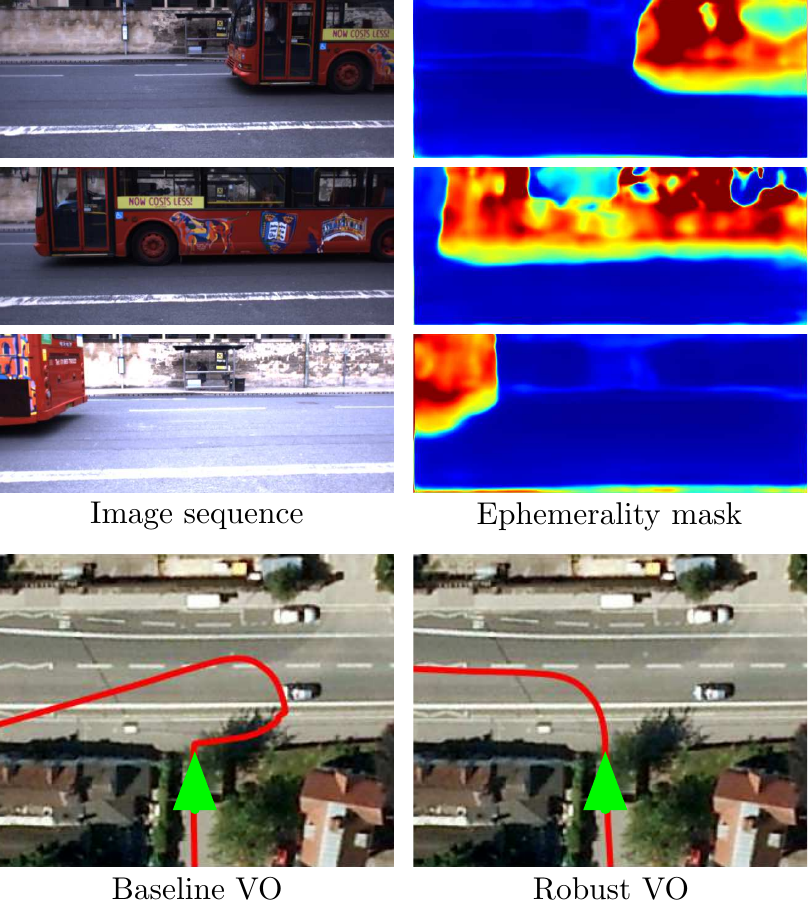}
  \caption{
  Robust motion estimation in urban environments using a single camera and a learned ephemerality mask. 
  When making a left turn onto a main road, a large bus passes in front of the vehicle (green arrow) obscuring the view of the scene (top left). 
  Our learned ephemerality mask correctly identifies the bus as an unreliable region of the image for the purposes of motion estimation (top right). 
  Traditional visual odometry (VO) approaches will incorrectly estimate a strong translational motion to the right due to the dominant motion of the bus (bottom left), whereas our approach correctly recovers the vehicle egomotion (bottom right).}
  \label{fig:front-ephemerality-mask}
%   \vspace{-4mm}
\end{figure}

In this paper we leverage large-scale offline mapping and deep learning approaches to produce a per-pixel \emph{ephemerality mask} at run-time without requiring any semantic classification or manual labelling, as illustrated in Fig. \ref{fig:front-ephemerality-mask} 
% and the accompanying video.
and the project video\footnotemark.
The ephemerality mask predicts stable image regions (e.g. buildings, road markings, static landmarks) that are likely to be useful for motion estimation, in contrast to dynamic or ephemeral objects (e.g. pedestrian and vehicle traffic, vegetation, temporary signage). 
In contrast to semantic segmentation approaches that explicitly label objects belonging to a-priori chosen classes and hence require manually annotated training data, our approach is trained using repeated traversals of the same route with a LIDAR-equipped survey vehicle producing per-pixel depth and ephemerality labels for a deep convolutional network as a fully self-supervised process.

We integrate the ephemerality mask as a component of a monocular visual odometry (VO) pipeline as an outlier rejection scheme. 
By leveraging the depth and ephemerality outputs of the network, we can produce robust metric-scale VO using \emph{only a single camera} mounted to a vehicle.
Our approach leads to significantly more reliable motion estimation when evaluated over hundreds of kilometres of driving in complex urban environments in the presence of heavy traffic and other challenging conditions.

\footnotetext{\color{blue}{\url{https://youtu.be/ebIrBn_nc-k}}}

%\begin{itemize}
%\item Visual navigation in dynamic urban environments is challenging \wmotes{TODO}
%\item In traffic, more than 50\% of an image can be occluded by moving objects \wmnotes{TODO}  
%\item Challenge for visual localisation and motion estimation \wmnotes{TODO} 
%\item We present a system that produces a per-pixel \emph{ephemerality mask} learned from prior data \wmnotes{TODO}
%\item We leverage large-scale mapping and offline processing to generate training data \wmnotes{TODO}
%\item Integrate the system with sparse and dense VO pipelines to produce robust monocular VO \wmnotes{TODO}
%\item Demonstrate reliable to-scale VO over 100s of km in urban environments \wmnotes{TODO}
%\item \textbf{Figure:} Ephemerality mask in dynamic environment \dbnotes{TODO}
%\end{itemize}

%%%%%%%%%%%%%%%%%%%%%%%%%%%%%%%%%%%%%%%%%%%%%%%%%%%%%%%%%%%%%%%%%%%%%%%%%%%%%%%%

\section{Related Work}\label{sec:related-work}

Estimating an ephemerality mask is closely related to background subtraction approaches \cite{piccardi2004background, jeeva2015survey}, which build statistics over background appearance based on training data from a static camera to identify discrepancies in live images.
These methods are typically used in surveillance applications and have limited robustness to general 3D camera motion in complex scenes, as experienced on a vehicle \cite{hayman2003statistical, sheikh2009background}.

Conversely, there is a significant body of work on detection and tracking of moving (foreground) objects \cite{yilmaz2006object, felzenszwalb2010object, ren2015faster}, which has been applied to robust VO in dynamic environments \cite{bak2014dynamic} and scale references for monocular SLAM \cite{song2014robust}.
However, these approaches require large quantities of manually-labelled training data of moving objects (e.g. cars, pedestrians) and the chosen object classes must cover all possibly-moving objects to avoid false negatives.
Recent 3D SLAM approaches have integrated per-pixel semantic segmentation layers to improve reconstruction quality \cite{civera2011towards, bowman2017probabilistic}, but again rely on laboriously manually-annotated training data and chosen classes that encompass all object categories.

Unsupervised approaches have recently been introduced to estimate depth \cite{garg2016unsupervised}, egomotion \cite{zhou2017unsupervised} and 3D reconstruction \cite{vijayanarasimhan2017sfm}. 
These methods are attractive for large-scale use as they only require raw video footage from a monocular or stereo camera, without any ground-truth motion estimates or semantic labels.
In particular, \cite{zhou2017unsupervised} introduces an ``explainability mask'', which highlights image regions that disagree with the dominant motion estimate.
However, the explainability mask differs from the ephemerality mask in that it only recognises non-dominant moving objects, and hence will still produce incorrect motion estimates when significantly occluded by a large, independently moving object.

Our approach is inspired by the distraction-suppression methods presented in \cite{mcmanus2013distraction, wolcott2016probabilistic}.
Both methods use a prior 3D map to estimate a mask that quantifies reliability for motion estimation, which is integrated into a VO pipeline.
We significantly extend the map prior approach of \cite{mcmanus2013distraction} to multi-session mapping and quantify ephemerality using a structural entropy metric, and use the result to automatically generate training data for a deep convolutional network.
As a result, our approach does not rely on live localisation against a prior map or live dense depth estimation from stereo, and hence can operate in a wider range of (unmapped) locations with a reduced (monocular-only) sensor suite.

%\begin{itemize}
%\item Classical background subtraction \wmnotes{TODO}
%\item Object detection and tracking \wmnotes{TODO}
%\item Semantic segmentation \wmnotes{TODO}
%\item Unsupervised Learning egomotion and depth - pitfalls \wmnotes{TODO}
%\item CNN-SLAM: metric VO from learned depth \wmnotes{TODO}
%\item Distraction suppression - no live localisation required \wmnotes{TODO}
%\end{itemize}

%%%%%%%%%%%%%%%%%%%%%%%%%%%%%%%%%%%%%%%%%%%%%%%%%%%%%%%%%%%%%%%%%%%%%%%%%%%%%%%%
% \vspace{-2mm}
\section{Learning Ephemerality Masks}\label{sec:overview}
% \vspace{-2mm}

In this section we outline our approach for automatically building ephemerality masks by leveraging an offline 3D mapping pipeline. 
Note that LIDAR and stereo camera sensors are only required for the survey vehicle to collect training data; at run-time only a monocular camera is required.
Our method takes the following steps:

\vspace{1mm} \noindent
\textbf{1) Prior 3D Mapping:} Using a survey vehicle equipped with a stereo camera and LIDAR scanner, we perform multiple traversals of the target environment. By analysing structural consistency across multiple mapping sessions with an entropy-based approach, we determine what constitutes the static (non-ephemeral) structure of the scene.

\vspace{1mm} \noindent
\textbf{2) Ephemerality Labelling:} We project the prior 3D static structure into every stereo camera image collected during the survey, and compare it to the structure computed by a dense stereo approach (similar to \cite{mcmanus2013distraction}). In the presence of traffic or dynamic objects these will differ considerably; we compute ephemerality as a weighted sum of disparity and normal difference between prior and true 3D structure. 

\vspace{1mm} \noindent
\textbf{3) Network Training:} We train a deep convolutional network to predict the resulting pixel-wise depth and ephemerality mask using only input monocular images. At run-time we produce live depth and ephemerality masks even in locations not traversed by the survey vehicle.

\vspace{1mm}
In the following sections we describe these steps in detail.

%\begin{itemize}
%\item \textbf{Figure:} System architecture \dbnotes{Need some kind of overview flow diagram (possibly hero fig) so show the total process}
%\end{itemize}

%%%%%%%%%%%%%%%%%%%%%%%%%%%%%%%%%%%
\subsection{Prior 3D Mapping}\label{sec:map-creation}

\begin{figure}[]
  \centering
  \includegraphics[width=1.0\columnwidth]{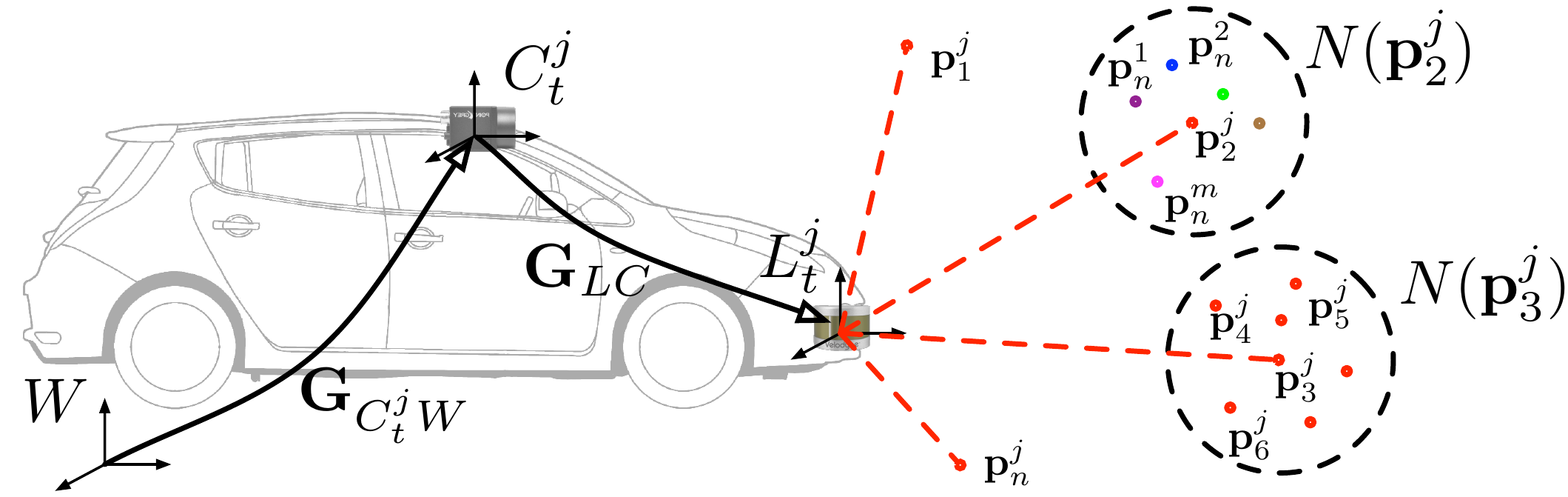}
  \caption{
%   \vspace{-1mm}
  Multi-session mapping and 3D pointcloud entropy computation. 
  For each traversal $j$, we compute the global pose of the vehicle $\mathbf{G}_{C_{t}^{j}W}$ at each timestamp $t$ and project points $\mathbf{p}$ into the global frame $W$. 
  We then analyse the neighbourhood $N$ of each point $\mathbf{p}_{i}^{j}$; in neighbourhoods where points are well distributed between traversals $\{1\cdots j\}$ such as $N(\mathbf{p}_{2}^{j})$ the scene is likely to be static, and where points are mostly derived from one traversal such as $N(\mathbf{p}_{3}^{j})$ the structure is ephemeral. 
  We quantify static scenes using an entropy metric applied to each neighbourhood $N(\mathbf{p})$.}
  \label{fig:lidar-diagram}
%   \vspace{-4mm}
\end{figure}

Given a survey vehicle equipped with a camera $C$ and LIDAR $L$ illustrated in Fig. \ref{fig:lidar-diagram} that has performed a number of traverses $j$ of an environment, we recover each global camera pose $\mathbf{G}_{C_{t}^{j}W}$ at time $t$ relative to world frame $W$ with a large-scale offline process using the stereo mapping and navigation approach in \cite{linegar2016made}.
We then compute the position of each 3D LIDAR point $\mathbf{p}_{i}^{j}\in\mathbb{R}^3$ in world frame $W$ using the camera pose and LIDAR-camera calibration $\mathbf{G}_{LC}$ as follows:

% \vspace{-1mm}
\begin{equation}\label{eqn:point-projection}
^{W}\mathbf{p}_{i}^{j} = \mathbf{G}_{C_{t}^{j}W}\mathbf{G}_{LC}\mathbf{p}_{i}^{j}
\end{equation}

\begin{figure*}[!htbp]
  \centering
  \includegraphics[width=1.0\textwidth]{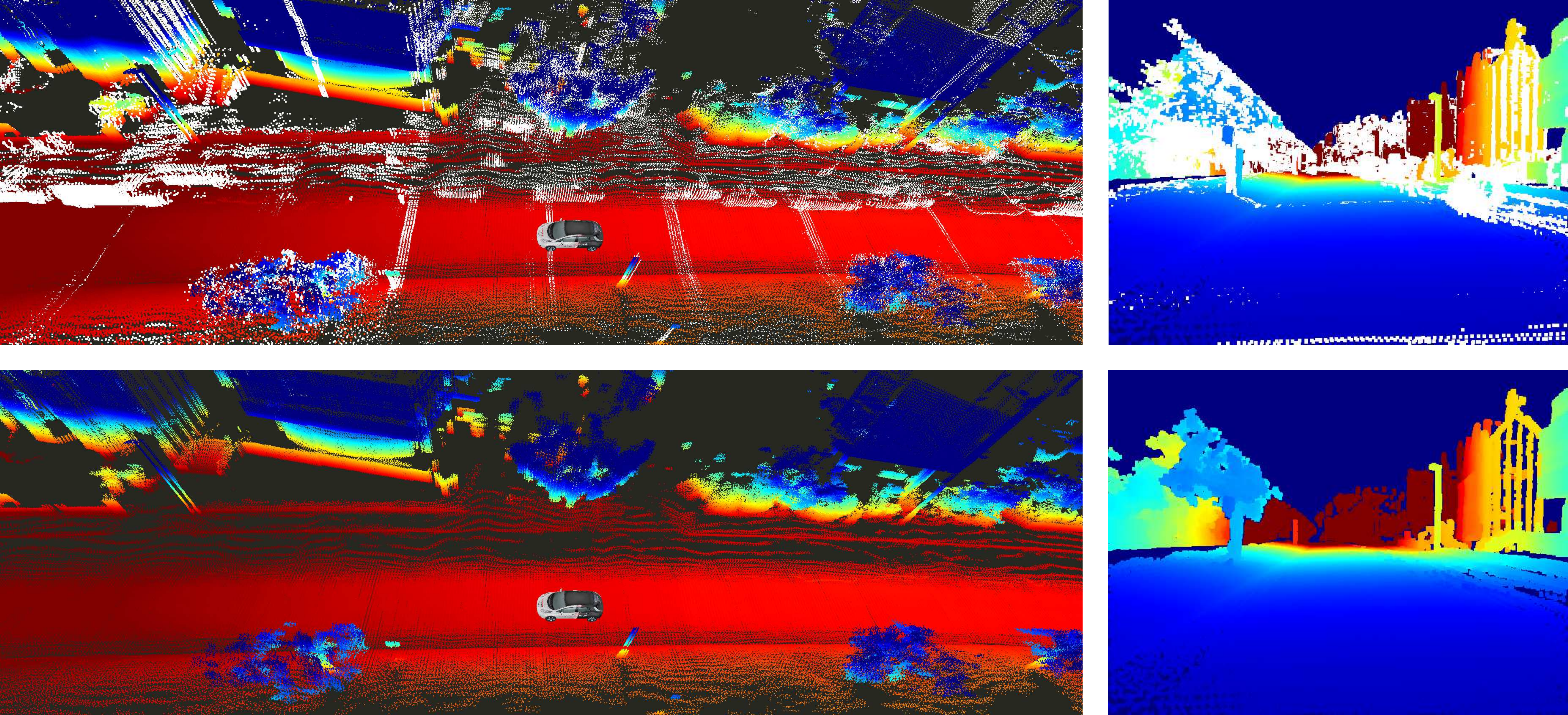}
  \caption{
  Prior 3D mapping to determine the static 3D scene structure. 
  Alignment of multiple traversals of a route (top left) will yield a large number of points only present in single traversals, e.g. traffic or parked vehicles, here shown in white.
  These points will corrupt a synthetic depth map (top right).
  Our entropy-based approach removes 3D points that were only observed in some traversals, and retains the structure that remained static for the duration of data collection (bottom left), resulting in high-quality synthetic depth maps (bottom right). 
  }
  \label{fig:dynamic-point-removal-diagram}
  \vspace{-4mm}
\end{figure*}

Given the pointcloud of all points $\mathbf{p}$ collected from all traversals $j$, we wish to compute the local entropy of each region of the pointcloud, to quantify how reliable the region is across each traversal.
We define a neighbourhood function $N(\cdot)$, where a point $\mathbf{p}_{t}^{k}$ belongs to a neighbourhood if it satisfies the following condition:

% \vspace{-1mm}
\begin{equation}\label{eqn:neighbourhood}
\mathbf{p}_{t}^{k} \in N(\mathbf{p}_{i}) \iff \left\Vert \mathbf{p}_{i} - \mathbf{p}_{t}^{k} \right\Vert _{2} < \alpha
\end{equation}

\noindent
where $\alpha$ is a neighbourhood size parameter, typically set to 0.5m in our experiments.
For each query point $\mathbf{p}_{i}$, we then build a distribution $p_{i}(j)$ over the traverses $j$ from which points fell in the neighbourhood of the query point as follows:

% \vspace{-1mm}
\begin{equation}\label{eqn:pdf}
p_{i}\left(j\right)=\dfrac{1}{\left|N\left(\mathbf{p}_{i}\right)\right|}\underset{\mathbf{p}_{t}^{k}\in N\left(\mathbf{p}_{i}\right)}{\sum}\left\{ \begin{array}{cc}
1, & j=k\\
0, & \textrm{otherwise}
\end{array}\right.
\end{equation}

Intuitively, neighbourhoods of points that are well-distributed between different traversals indicate static structure, whereas neighbourhoods of points that were only sourced from one or two traversals are likely to be ephemeral objects.
We compute the neighbourhood entropy $\textrm{H}(p_i)$ of each point across all $n$ traversals as follows:

\begin{equation}\label{eqn:entropy}
\textrm{H}\left(p_{i}\right)=-\underset{j=1}{\overset{n}{\sum}}p_{i}\left(j\right)\log\left(p_{i}\left(j\right)\right)  
\end{equation}

We classify a point $\mathbf{p}_i$ as static structure $\mathbf{p}_i^S$ if the neighbourhood entropy $\textrm{H}(p_{i})$ exceeds a minimum threshold $\beta$; all other points are estimated to be ephemeral and are removed from the static 3D prior. 
The pointcloud construction, neighbourhood entropy and ephemeral point removal process are illustrated in Fig. \ref{fig:dynamic-point-removal-diagram}.

%\begin{itemize}
%\item Multi-session map alignment \wmnotes{TODO}
%\item Local LIDAR projection \wmnotes{TODO}
%\item Per-point entropy computation \wmnotes{TODO}
%\item \textbf{Figure:} Sensor and neighbourhood diagram \wmnotes{TODO}
%\item \textbf{Figure:} 3D map alignment \dbnotes{TODO}
%\item \textbf{Figure:} Dynamic point removal from 3D map \dbnotes{TODO}
%\end{itemize}

%%%%%%%%%%%%%%%%%%%%%%%%%%%%%%%%%%%
\subsection{Ephemerality Labelling}\label{sec:ephemerality-labelling}

\begin{figure*}[h!t]
  \centering
  \includegraphics[width=1.0\textwidth]{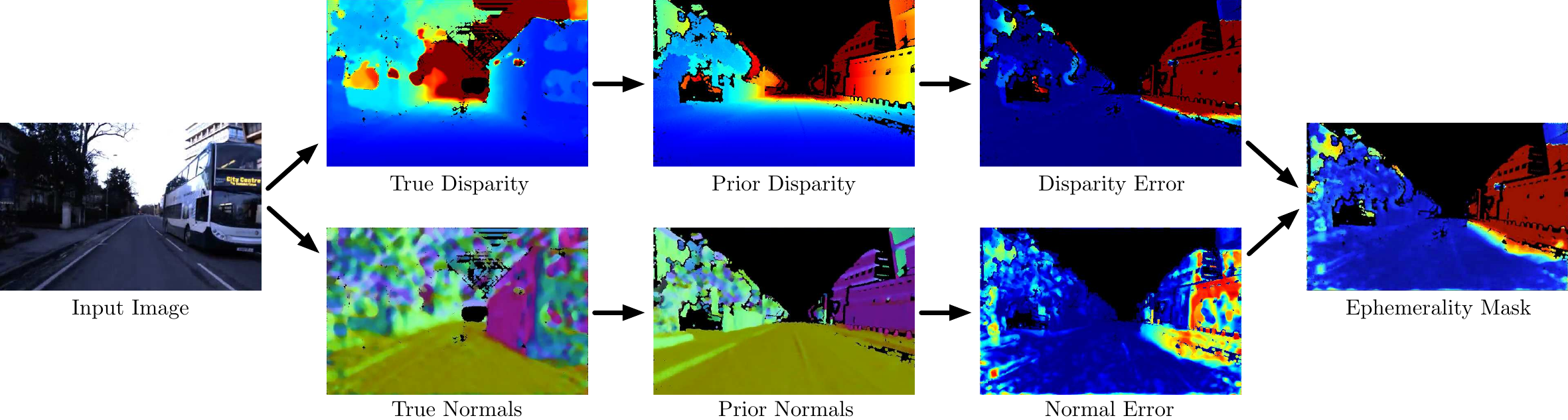}
  
\vspace{-2mm}
  \caption{
  Ephemerality labelling process. 
  From input images (left) we compute the true disparity $d_i$ and normals $\mathbf{n}_i$ using an offline dense stereo approach. 
  We then project the prior 3D pointcloud $\mathbf{p}^S$ into the image to form the prior disparity $d_i^S$ and prior normal $\mathbf{n}_i^S$. 
  The disparity and normal error terms are combined to form the ephemerality mask (right). 
  }
  \label{fig:ephemerality-diagram}
  
  \vspace{-3mm}
\end{figure*}

Given the prior 3D static pointcloud $\mathbf{p}^S$ and globally aligned camera poses $C$, we can produce a synthetic depth map for each survey image, as illustrated in Fig. \ref{fig:ephemerality-diagram}.
To handle visibility constraints we make use of the hidden point removal approach in \cite{katz2007direct}.
For every pixel $i$ into which a valid prior 3D point projects, we compute the expected disparity $d_{i}^S$ and normal $\mathbf{n}_{i}^S$ using the local 3D structure of the pointcloud.

In the presence of dynamic objects, the scene observed from the camera will differ from the expected prior 3D map.
We use the offline dense stereo reconstruction approach of \cite{hirschmuller2005accurate} to compute the true disparity $d_{i}$ and normal $\mathbf{n}_{i}$ for each pixel in the survey image, illustrated in Fig. \ref{fig:ephemerality-diagram}.
We define the ephemerality mask $\mathcal{E}_{i}$ as the weighted difference between the expected static and true disparity and normals as follows:

\begin{equation}\label{eqn:ephemerality}
\mathcal{E}_{i}=\gamma\left\Vert d_{i}^{S}-d_{i}\right\Vert _{1}+\delta\cos^{-1}\left(\mathbf{n}_{i}^{S}\cdot\mathbf{n}_{i}\right)
\end{equation}

\noindent
where $\gamma$ and $\delta$ are weighting parameters, and $\mathcal{E}_{i}$ is bounded to $[0, 1]$ after computation.

%\begin{itemize}
%\item 3D to 2D map projection
%\item Disparity and normal computation from prior
%\item Live disparity and normal computation
%\item Ephemerality mask weighting (depth + normal)
%\item \textbf{Figure:} 3D prior map projection, normal computation, live depth and normal
%\item \textbf{Figure:} Disparity and normal error term, combined mask
%\end{itemize}

\begin{figure}[]
  \centering
  \includegraphics[width=1.0\columnwidth]{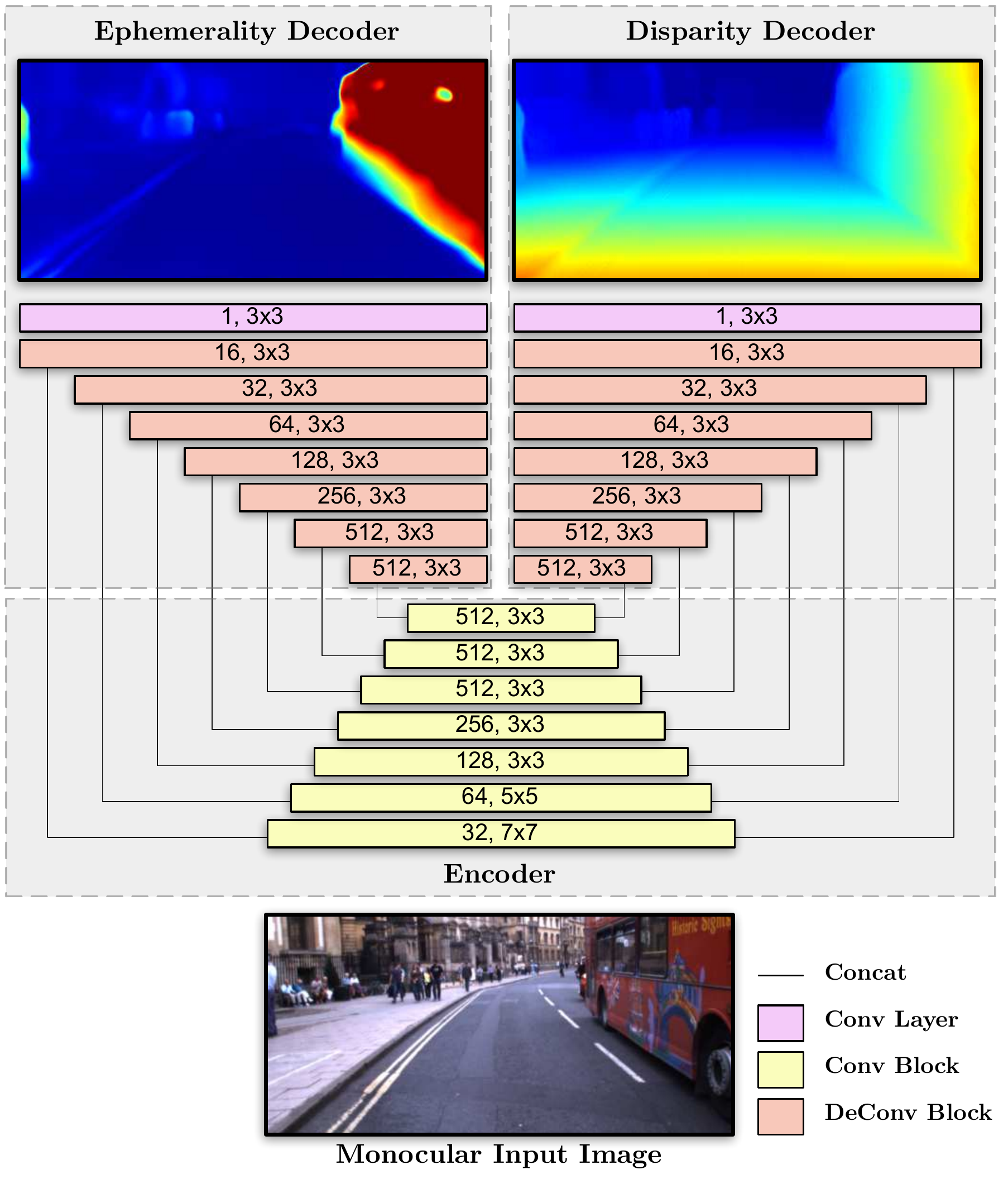}
  
\vspace{-2mm}
  \caption{
  Network architecture for ephemerality and disparity \mbox{learning}. 
  The width of each block indicates the spatial dimensions of the feature map, which vary by a factor of 2 between blocks. The number of output channels and filter dimensions are also detailed for each block. 
  }
  \label{fig:network-diagram}
\vspace{-2mm}
%   \vspace{-5mm}
\end{figure}

%%%%%%%%%%%%%%%%%%%%%%%%%%%%%%%%%%%
\subsection{Network Architecture}\label{sec:network-architecture}

% Comments: Changed Network Architecture Text
% Comments: Removed UNet sentence
We adopt a convolutional encoder-multi-decoder network architecture to predict both disparity and ephemerality masks from a single image, as illustrated in Fig. \ref{fig:network-diagram}, by adding an additional decoder to the architecture in \cite{monodepth17}.
% The skip connections between the encoder and decoder layers help preserve fine structure similar to the U-Net approach \cite{ronneberger2015u}.

To train the disparity output we use the stereo photometric loss proposed in \cite{monodepth17}, optionally semi-supervised using the prior LIDAR disparity $d_i^S$ to ensure metric-scaled outputs. For the ephemerality output we use the $\mathcal{L}_1$ loss for each pixel with a valid ephemerality label.
We balance these losses using the multi-task learning approach in \cite{kendall2017multi}, which continuously updates the inter-task weighting during training. 

As in \cite{monodepth17}, we trained our model from scratch for 50 epochs, with a batch size of 8 using the Adam \cite{kingma2014adam} optimiser, with $\beta_1$ = 0.9, $\beta_2$ = 0.999, and $\epsilon$ = $10^{-8}$. We used an initial learning rate of $\lambda$ = $10^{-4}$ which we kept constant for the first 30 epochs before halving it every 10 epochs until the
end.

%\begin{itemize}
%\item Training data generation, inputs and outputs
%\item Photometric depth error, optionally supervised depth error
%\item Architecture, losses and parameters
%\item \textbf{Figure:} Diagram of network structure, layers and inputs/outputs
%\end{itemize}

% Comments: Changed Network Architecture Diagram

% \begin{figure}[]
%   \centering
%   \includegraphics[width=1.0\columnwidth]{figs/other/NetworkICRA_ver3_crop.pdf}
%   \caption{
%   Network architecture for ephemerality and disparity learning. 
%   We use a common encoder which splits to multiple decoders for the ephemerality mask and disparity outputs. 
%   }
%   \label{fig:network-diagram-old}
% %   \vspace{-2mm}
% \end{figure}

%%%%%%%%%%%%%%%%%%%%%%%%%%%%%%%%%%%%%%%%%%%%%%%%%%%%%%%%%%%%%%%%%%%%%%%%%%%%%%%%

\section{Ephemerality-Aware Visual Odometry}\label{sec:robust-vo}

We leverage the live depth and ephemerality mask produced by the network to produce reliable visual odometry estimates accurate to metric scale.
We present two robust VO approaches: a sparse feature-based approach and a dense photometric approach.
Each integrates the ephemerality mask in order to estimate egomotion using only static parts of the scene, and uses the learned depth to estimate relative motion to the correct scale.
This improves upon traditional monocular VO systems that cannot recover absolute scale \cite{strasdat2010scale}.
Both our odometry approaches are optimised for real-time performance on a vehicle platform.

\begin{figure}[]
  \centering
  \includegraphics[width=1.0\columnwidth]{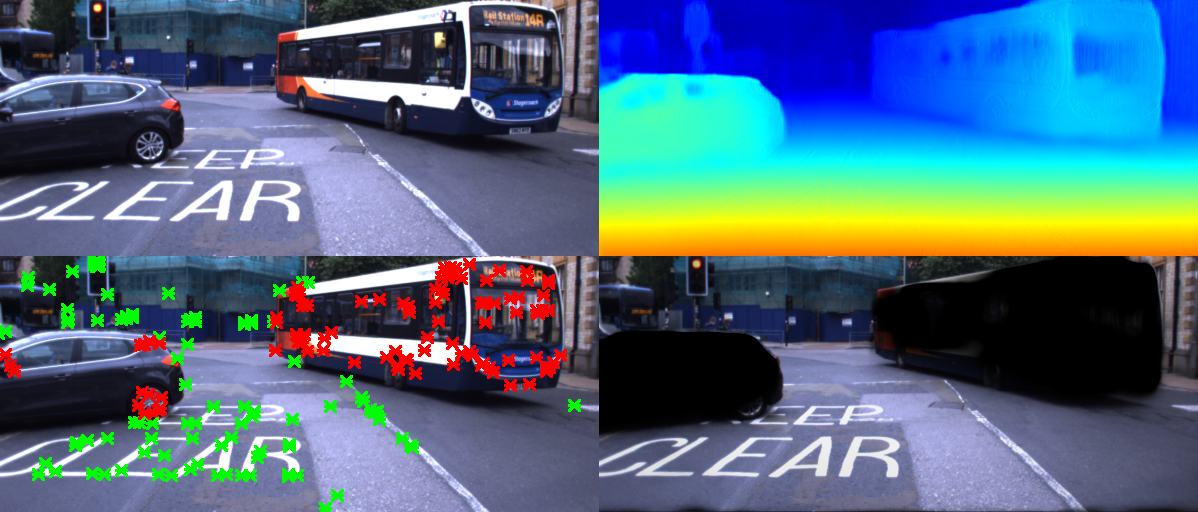}
  
\vspace{-2mm}
  \caption{Input data for ephemerality-aware visual odometry. For a given input image (top left), the network predicts a dense depth map (top right) and an ephemerality mask. For sparse VO approaches, the ephemerality mask is used to select which features are used for optimisation (bottom left), and for dense VO approaches the photometric error term is weighted directly by the ephemerality mask (bottom right).}
  \label{fig:vo-feature-vis}
%   \vspace{-8mm} 
\end{figure}

%\begin{itemize}
%\item Overview of three methods
%\item \textbf{Figure:} All points, robust points only and robust depths only
%\end{itemize}

%%%%%%%%%%%%%%%%%%%%%%%%%%%%%%%%%%%
\subsection{Sparse Monocular Odometry}\label{sec:mono-sparse-vo}

Our sparse monocular VO approach is derived from well-known stereo approaches \cite{nister2006visual}, where sets of features are detected and matched across successive frames to build a relative pose estimate.
Each feature $\mathbf{x}_i$ is parameterised as follows:

\vspace{-2mm}
\begin{equation}
\mathbf{x}_{i}=\left[\begin{array}{c}
u_{i}\\
v_{i}\\
d_{i}
\end{array}\right]
\end{equation}

\noindent
where $(u_i,v_i)$ are the pixel coordinates and $d_i$ is the disparity predicted by the deep convolutional network.
The relative pose $\boldsymbol{\xi}\in\mathbb{SE}(3)$ is recovered by minimising the reprojection error between matched features $\mathbf{x}_i$ and $\mathbf{\hat{x}}_i$: 

\begin{equation}
\underset{\boldsymbol{\xi}}{\arg\min}\underset{i\in\mathcal{F}}{\sum}s\left(\mathcal{E}_{i}\right)\left\Vert \mathbf{x}_{i}-\omega\left(\mathbf{\hat{x}}_{i},\boldsymbol{\xi}\right)\right\Vert _{2}^{2}
\end{equation}

The warping function $\omega(\cdot)\rightarrow\mathbb{R}^{2}$ projects the matched feature $\mathbf{\hat{x}}_i$ into the current image according to relative pose $\boldsymbol{\xi}$ and the camera intrinsics.
The set of all extracted features $\mathcal{F}$ is typically a small subset of the total number of pixels in the image.
The step function $s(\mathcal{E}_i)$ is used to disable the residual according to the predicted ephemerality as follows:

\begin{equation}
s\left(\mathcal{E}_{i}\right)=\left\{ \begin{array}{cc}
1, & \mathcal{E}_{i}<\tau\\
0, & \textrm{otherwise}
\end{array}\right.
\end{equation}

\noindent
where $\tau$ is the maximum ephemerality threshold for a valid feature, typically set to 0.5.
In practice we detect sparse features using FAST corners \cite{rosten2006machine} and match using BRIEF descriptors \cite{calonder2010brief} for real-time operation.

%%%%%%%%%%%%%%%%%%%%%%%%%%%%%%%%%%%
\subsection{Dense Monocular Odometry}\label{sec:mono-dense-vo}

For the dense monocular approach, we adopt the method of \cite{tateno2017cnn} and combine our learned depth maps with the photometric relative pose estimation of \cite{engel2013semi}.
Rather than a subset of pixels $\mathcal{F}$, all pixels $i$ within the reference keyframe image $\mathcal{I}_r$ are warped into the current image $\mathcal{I}_c$ and the relative pose $\boldsymbol{\xi}$ is recovered by minimising the photometric error as follows:

\begin{equation}
\underset{\boldsymbol{\xi}}{\arg\min}\underset{i\in\mathcal{I}_{r}}{\sum}\left(1-\mathcal{E}_{i}\right)\left\Vert \mathcal{I}_{r}\left(\mathbf{x}_{i}\right)-\mathcal{I}_{c}\left(\omega\left(\mathbf{x}_{i},\boldsymbol{\xi}\right)\right)\right\Vert _{2}^{2}
\end{equation}

\noindent
where the image function $\mathcal{I}(\mathbf{x}_i)\rightarrow\mathbb{R}^{+}$ returns the pixel intensity at location $(u_i,v_i)$. 
Note that the ephemerality mask is used directly to weight the photometric residual; no thresholding is required.
Fig. \ref{fig:vo-feature-vis} illustrates the predicted depth, selected sparse features and weighted dense intensity values used for a typical urban scene.

%%%%%%%%%%%%%%%%%%%%%%%%%%%%%%%%%%%%%%%%%%%%%%%%%%%%%%%%%%%%%%%%%%%%%%%%%%%%%%%%

\section{Experimental Setup}\label{sec:experimental-setup}
\vspace{-1mm}
We benchmarked our approach using hundreds of kilometres of data collected from an autonomous vehicle platform in a complex urban environment.
Our goal was to quantify the performance of the ephemerality-aware visual odometry approach in the presence of large dynamic objects in traffic.

%%%%%%%%%%%%%%%%%%%%%%%%%%%%%%%%%%%
\subsection{Network Training}\label{sec:stereo-vo}
\vspace{-1mm}
We train our approach using eight 10km traversals from the Oxford RobotCar dataset \cite{RobotCarDatasetIJRR} for a total of approximately 80km of driving.
The RobotCar vehicle is equipped with a Bumblebee XB3 stereo camera and a LMS-151 pushbroom LIDAR scanner.
For training we downsample the input images to $640\times256$ pixels and subsample to one image every metre before use; a total of 60,850 images were used for training. 
At run-time we produce ephemerality masks and depth maps at $50$Hz using a single GTX 1080 Ti GPU.

%\begin{itemize}
%\item Training routes, alignment with Dub4
%\item Test routes and test conditions
%\end{itemize}

%%%%%%%%%%%%%%%%%%%%%%%%%%%%%%%%%%%
\subsection{Evaluation Metrics}\label{sec:mono-sparse-vo}
\vspace{-1mm}
% Comments: Altered first sentence to gain line
We evaluate our approach on 42 further Oxford traversals for a total of over 400km.
The evaluation datasets contain multiple detours and alternate routes, ensuring the method is tested in (unmapped) locations not present in the training datasets.
To quantify the performance of the ephemerality-aware VO, we compute translational and rotational drift rates using the approach proposed in the KITTI odometry benchmark \cite{geiger2012we}.
Specifically, we compute the average end-point-error for all subsequences of length $(100,200,\ldots,800)$ metres compared to the INS system installed on the vehicle.

In addition, we compare the instantaneous translational velocities of each method to that reported by the INS system (based on doppler velocity measurements).
We manually selected 6,000 locations that include distractors, and evaluate velocity estimation errors in comparison to the average of all locations.
This allows us to focus on dynamic scenes where independently moving objects produce erroneous velocity estimates in the baseline VO methods.

%\begin{itemize}
%\item Evaluation vs INS system
%\item Average end-point-error (drift rates)
%\item Instantaneous velocity
%\end{itemize}

%%%%%%%%%%%%%%%%%%%%%%%%%%%%%%%%%%%%%%%%%%%%%%%%%%%%%%%%%%%%%%%%%%%%%%%%%%%%%%%%

% Comments: Made non line breaking
\hyphenpenalty=10000
  
\section{Results}\label{sec:results}
\begin{figure*}[h!t]
  \centering
  \includegraphics[width=1.0\textwidth]{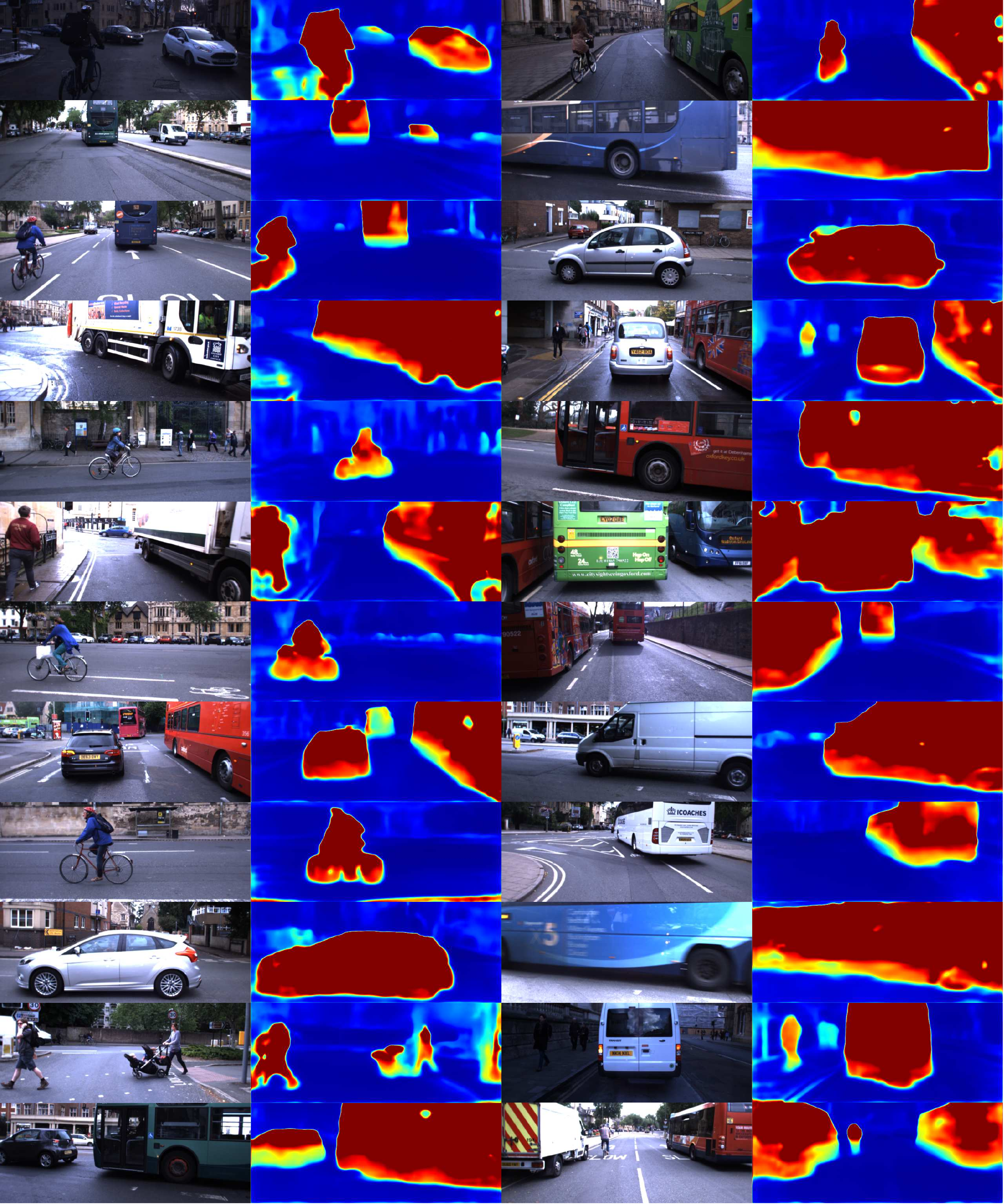}
  \caption{
  Ephemerality masks produced in challenging urban environments.
  The masks reliably highlight a diverse range of dynamic objects (cars, buses, trucks, cyclists, pedestrians, strollers) with highly varied distances and orientations.
  Even buses and trucks that almost entirely obscure the camera image are successfully masked despite the lack of other scene context.
  Robust VO approaches that make use of the ephemerality mask can provide correct motion estimates even when more than 90\% of the static scene is occluded by an independently moving object.
  }
  \label{fig:qualitative-images}
\end{figure*}

\vspace{-1mm}
In addition to the quantitative results listed below, we present qualitative results for ephemerality masks produced in a range of different locations in Fig. \ref{fig:qualitative-images}.

\subsection{Odometry Drift Rates}
\vspace{-1mm}
The end-point-error evaluation for each of the methods is presented in Table \ref{tab:kitt-style-vo-results}.
In both cases, the addition of the ephemerality mask reduced both average translational and rotational drift over the full set of evaluation datasets.
Note that the metric scale for translational drift is derived from the depth map produced by the network, and hence both systems report translation in units of metres with low overall error rates using only a monocular camera.
The sparse VO approach provided lower overall translational drift, whereas the dense approach produced lower orientation drift. 

\subsection{Velocity Estimates}
\begin{figure}[]
  \centering
  \includegraphics[width=1.0\columnwidth]{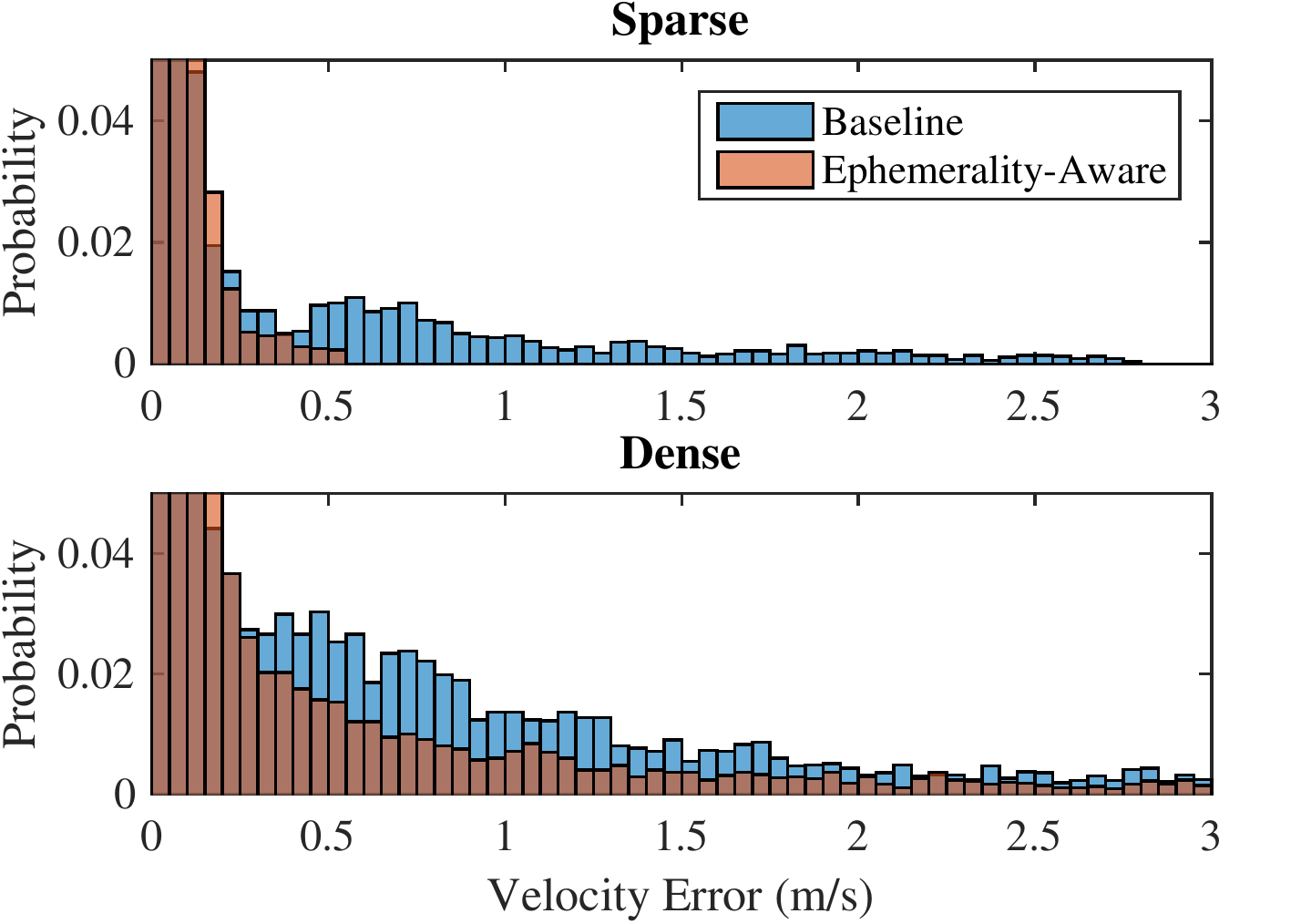}
  \vspace{-6mm}
  \caption{Velocity estimation errors in the presence of distractors. 
  The sparse ephemerality-aware approach significantly outperforms the baseline approach, producing far fewer outliers above 0.5 m/s.
  The dense ephemerality-aware approach does not perform as well, but still outperforms the baseline.
  The vertical axis is scaled to highlight the outliers.}
  \label{fig:distractor-histogram}
\end{figure}

\vspace{-1mm}
The velocity error evaluation for each of the methods is presented in Table \ref{tab:velocity-errors}.
Across all the evaluation datasets, the ephemerality-aware odometry approaches produce lower average velocity errors.
However, in locations with distractors, the ephemerality-aware approaches produce significantly more accurate velocity estimates than the baseline approaches.
In particular, the robust sparse VO approach is almost unaffected by distractors, whereas the baseline method reports errors 4 times greater. 
The dense VO approach generally produces poorer translational velocity estimates than the sparse approach, which corresponds with higher translational drift rates reported in the previous section.
Fig. \ref{fig:distractor-histogram} presents the distribution of velocity errors for each of the approaches in the presence of distractors.

%\begin{itemize}
%\item Qualitative ephemerality masks in traffic, rain, overexposure
%\item Quantitative VO drift and velocity results
%\item Compute time
%\item \textbf{Figure:} Qualitative ephemerality masks
%\item \textbf{Table:} End-point errors for methods
%\item \textbf{Figure:} Graph or histogram of instantaneous velocity errors
%\item \textbf{Figure:} 3D reconstruction of location with strong dynamic motion (e.g. bus)
%\end{itemize}

\begin{table}
\centering
\caption{Odometry Drift Evaluation}
\vspace{-2mm}
\begin{tabular}{C{3cm} | C{1.5cm} | C{2.5cm} }
 \textbf{VO Method} & \textbf{Translation [\%]} & \textbf{Rotation [deg/m]} \\
 \hline
% Sparse Stereo Normal       & 6.95              & 0.0336 &
% Sparse Stereo Distraction  & \textbf{6.67}     & \textbf{0.0305} & \hline
 Sparse & 6.55              & 0.0353 \\
 Sparse w/Ephemerality    & \textbf{6.38}     & \textbf{0.0321} \\ \hline
 Dense & 7.15              & 0.0373 \\
 Dense w/Ephemerality & \textbf{6.52}     & \textbf{0.0307} 
\end{tabular}
\label{tab:kitt-style-vo-results}
\vspace{-2mm}
\end{table}

\begin{table}
\centering
\caption{Velocity Error Evaluation}
\vspace{-2mm}
\begin{tabular}{C{3cm} | C{1.5cm} | C{2.5cm} }
 \textbf{VO Method} & \textbf{All [m/s]} & \textbf{Distractors [m/s]} \\
 \hline
 Sparse                     & 0.0548              & 0.220 \\
 Sparse w/Ephemerality      & \textbf{0.0406}     & \textbf{0.0489} \\ \hline
 Dense                      & 0.0568              & 0.766 \\
 Dense w/Ephemerality       & \textbf{0.0407}     & 0.424
\end{tabular}
\label{tab:velocity-errors}
\vspace{-6mm}
\end{table}

%%%%%%%%%%%%%%%%%%%%%%%%%%%%%%%%%%%%%%%%%%%%%%%%%%%%%%%%%%%%%%%%%%%%%%%%%%%%%%%%

\section{Conclusions}\label{sec:conclusions}
\vspace{-1mm}
%\wmnotes{Introductory sentence}
In this paper we introduced the concept of an ephemerality mask, which estimates the likelihood that any pixel in an input image corresponds to either reliable static structure or dynamic objects in the environment, and can be learned using an automatic self-supervised approach.
Crucially, we do not require any manual labelling or choice of semantic classes in order to train our approach, and at run-time we only require a single monocular camera to produce reliable ephemerality-aware visual odometry to metric scale.
Over hundreds of kilometres our approach produces improved odometry resulting in lower drift rates, and significantly more robust velocity estimates in the presence of large dynamic objects in urban scenes.

\begin{figure}[]
  \centering
  \includegraphics[width=1.0\columnwidth]{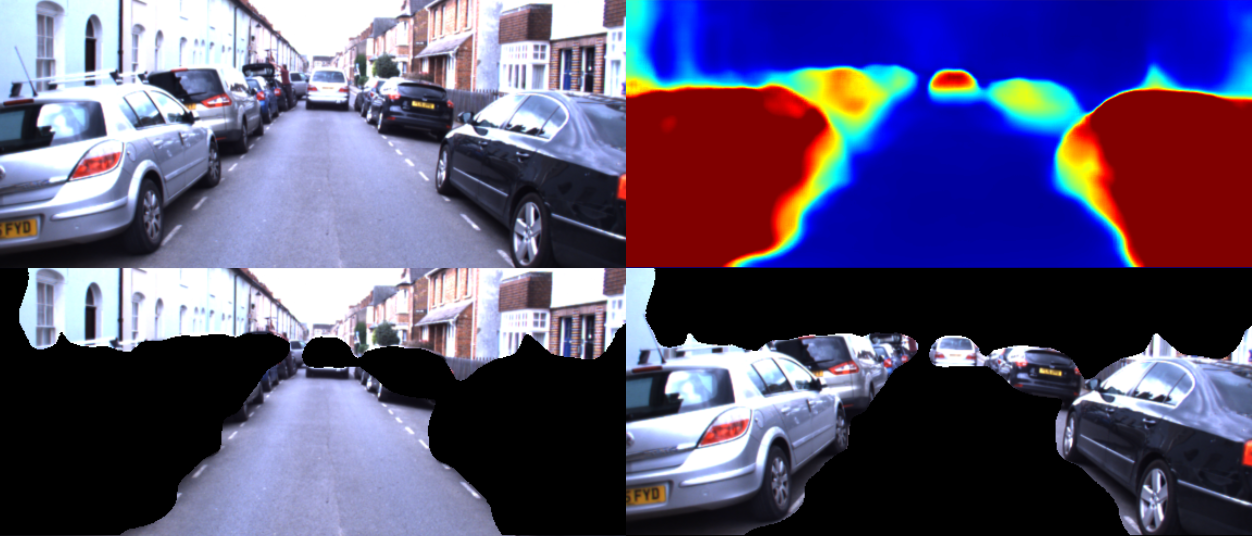}
  \vspace{-4mm}
  \caption{Ephemerality masks are widely applicable for autonomous vehicles. In the above scene the ephemerality mask can be used to inform localisation against only the static scene (bottom left) whilst guiding object detection to only the ephemeral elements (bottom right). }
  \label{fig:conclusion-foreground-background}
  \vspace{-4mm}
\end{figure}

The benefits of our approach are not restricted to improving motion estimation, and there are a number of avenues to explore in future work. 
Fig. \ref{fig:conclusion-foreground-background} illustrates a foreground/background segmentation performed using the ephemerality mask; where we currently use the background to guide motion estimation, a detection and classification approach could be guided by the foreground mask to efficiently track dynamic objects in the scene.
We plan to integrate the approaches in this paper for improved localisation, motion estimation, obstacle avoidance and scene understanding for fully autonomous vehicles operating in complex urban \mbox{environments}.

% Comments: changed from this sentence for space
%  a detection and classification approach could be guided by the foreground mask to efficiently track dynamic objects in the scene.

%\begin{itemize}
%\item Reliable motion estimats in 6DoF from a single camera in dynamic urban environments
%\item Introduced concept of ephemerality mask, applied to sparse and dense VO
%\item Extend to localisation and object detection
%\item \textbf{Figure:} Background and foreground masks from ephemerality
%\end{itemize}

%%%%%%%%%%%%%%%%%%%%%%%%%%%%%%%%%%%%%%%%%%%%%%%%%%%%%%%%%%%%%%%%%%%%%%%%%%%%%%%%
\section{Acknowledgements}\label{sec:conclusions}

This work was supported by the UK EPSRC Doctoral Training Partnership and Programme Grant EP/M019918/1.

\bibliographystyle{IEEEtran}
\bibliography{icra2018}

\end{document}